\documentclass[smallcondensed]{llncs} 
\pdfoutput=1
\usepackage{graphicx}
\usepackage{color}

\newcommand{\minimaloptimal}{\textit{minimal optimal} }
\newcommand{\allrelevant}{\textit{all relevant} }
\newcommand{\code}[1]{\texttt{#1}}
\newcommand{\rf}{random forest }
\newcommand{\pkgname}[1]{\textit{#1}}
\newcommand{\FP}{\mathit{FP}}
\newcommand{\TP}{\mathit{TP}}
\newcommand{\FN}{\mathit{FN}}

\newcommand{\Prec}{\mathit{Prec}}
\newcommand{\Rec}{\mathit{Rec}}

\begin{document}

\title{The All Relevant Feature Selection using Random Forest }

\author{Miron B. Kursa\inst{1}  \and Witold R. Rudnicki\inst{1}}

\title{The All Relevant Feature Selection using Random Forest }

\institute{Interdisciplinary Centre for Mathematical and Computational Modelling\\
       University of Warsaw\\
       Pawinskiego 5a, 02-106 Warsaw\\
              \email{ \{M.Kursa,W.Rudnicki\}@icm.edu.pl}}

\maketitle

\begin{abstract}
In this paper we examine the application of the random forest classifier for the all relevant feature selection problem.  
To this end we first examine two recently proposed all relevant feature selection algorithms, both being a random forest wrappers, on a series of synthetic data sets with varying size.
We show that reasonable accuracy of predictions can be achieved and that heuristic algorithms that were designed to handle the all relevant problem, have performance that is close to that of the reference ideal algorithm.  

Then, we apply one of the algorithms to four families of semi-synthetic data sets to assess how the properties of particular data set influence results of feature selection. 

Finally we test the procedure using a well-known gene expression data set. 
The relevance of  nearly all previously established important genes was confirmed, moreover the relevance of several new ones is discovered. 
\end{abstract}

\section{Introduction}
Feature selection is indenspensable stage of data analysis when dealing with datasets described with thousands variables. 
It is applied to select a much smaller set of features that are \emph{relevant} for classification. 
Historically the relevance of the variables has been defined in many different ways that were not compatible with each other, as demonstrated in \cite{Kohavi1997b}.  
This happened because various researchers concentrated on different notions that can be associated with this term. 
Kohavi \cite{Kohavi1997b} proposed to use two degrees of relevance (\emph{strong} and \emph{weak}) to encompass all these notions. 
In their approach the relevance is defined in the absolute terms, with the help of ideal Bayes classifier. 
Feature $X$ is defined to be \emph{strongly relevant} when removal of $X$ alone from the data always results in deterioration of the prediction accuracy of the ideal Bayes classifier. 
Feature $X$ is \emph{weakly relevant} if it is not strongly relevant and there exists a subset of features, $S$, such that the performance of ideal Bayes classifier on $S$ is worse than  the performance on $S\cup\{X\}$.
A feature is \emph{irrelevant} if it is neither strongly nor weakly relevant.

One should note that an information system can be constructed so that it contains no strongly relevant attributes; for example the Madelon set used in the NIPS 2003 feature selection contest \cite{Guyon2005} consists of points in five-dimensional space that are described with twenty relevant attributes along with several hundred irrelevant ones. 
The set of relevant attributes consists of five original point coordinates as well as fifteen linear combinations of the basic attributes, hence all these attributes are weakly relevant. 

\subsection{Two feature selection problems}
The feature selection problem arose due to practical reasons --- dealing with small sets of relevant attributes is easier and usually gives better results \cite{Ambroise2002d,Hua2005b,Janecek2008,Hua2009a}.  
Nevertheless, in many applications a robust feature selection might be a more interesting problem than a simple reduction of the processing time or even improving classification accuracy. 
Taking an example from biology, finding a minimal set that is optimal for the classification task is useful for designing the diagnostic test, but brings no thorough understanding of the process.
On the other hand, finding all relevant features (such as genes relevant for a particular type of cancer) might help researchers to decipher mechanisms underlying problems of interest (cancerogenesis), \cite{Jirapech-Umpai2005c,Draminski2008}.
In that context the problem often described as finding all differentially expressed genes and is explored using statistical tools \cite{Golub1999h,Dudoit2002h,Dudoit2003c,Nilsson2007}. 

Recently, Nilsson \cite{Nilsson2007} proposed to define formally two distinct problems in feature selection: 
\begin{itemize}
\item The \minimaloptimal problem is finding a set consisting of all strongly relevant attributes and such subset of weakly relevant attributes that all remaining weakly relevant attributes contain only redundant information. 
\item The \allrelevant problem is finding all strongly and weakly relevant attributes. 
\end{itemize}
It has been shown there that the exact solution of the \allrelevant problem requires an exhaustive search, and is thus realiseable only for a smallest systems. 

\subsection{Methods for feature selection}
There are three general classes of feature selection algorithms --- filters, wrappers and embedded algorithms \cite{Guyon2003b}. 

Filters are based on some importance measure which is independent from any classification method, such as for example correlation between features and decision class \cite{Miyahara2000} or information gain \cite{Janecek2008}. 
They are applied before the classification task is performed. 
Unfortunately filter methods are not designed to detect complex relations between features and  with the decision and generally are not sensitive enough for the \allrelevant problem.  

Embedded algorithms perform feature selection during the classifier training procedure, namely they explicitly optimise the attribute set used to achieve best accuracy.
Hence they solve the \minimaloptimal problem.  
We are not aware of any embedded method designed to solve the \allrelevant problem.

Finally the  wrapper methods rely on the information about feature relevance obtained from some classification method \cite{Kohavi1997b} and therefore may use deeper insight in data than filters. 
Wrappers are usually created around particular classifier, but may in principle use any classifier that also provides some measure of feature importance.
The wrapper methods are best suited for \allrelevant feature selection; the computational effort they require is usually significant, but inevitable to get comprehensive view in attribute relevance. 

\subsection{\textit{All relevant} feature selection}
There are two issues that are central for the \allrelevant problem but non-existent for the  \minimaloptimal one. 
The first issue is detection of weakly relevant attributes that can be completely obscured by other attributes, the second is discerning between weakly but truly relevant variables from those that are only seemingly relevant due to random fluctuations.

The concepts of strong and weak relevance, and consequently also a problem of \allrelevant feature selection, are defined in a context of a perfect classifier that is able to use all available information.  
Yet, in the real world applications one is restricted to use imperfect classification algorithms that are not capable to use all information present in the information system, what may influence the outcome of the feature selection algorithm. 
In particular, an algorithm may not be able to find and use some of the relevant features.

In many cases this won't disturb the solution of a \minimaloptimal problem, provided that the final predictions of the classifier are sufficiently accurate; yet it will significantly decrease the sensitivity of \allrelevant feature selection. 
Hence, the classification algorithm used in \allrelevant feature selection should be able to detect weak and redundant attributes.

The features of the \rf algorithm \cite{Breiman2001} make it a promising candidate for this task. 
It is an ensemble of numerous weak classifiers (decision trees), each of these classifiers is constructed using different subset of variables and different subset of objects.  
During the construction process, each variable has numerous chances to be included in the classifier, so even weakly relevant attributes that are marginally related with the decision attribute will be used for construction of individual classifiers. 
Moreover, the importance of the attribute is measured using only the trees that use given attribute, therefore the signal for attributes contributing to a small number of trees is still visible.
In addition the \rf has some additional advantages for use as wrapper's engine; it has very few tunable parameters, is relatively efficient computationally and the importance evaluation does not incur significant computational overheads. 

The second issue, namely discerning between the truly and randomly relevant attributes arises because the analyses are performed for a finite size samples.
This gives a chance for random correlations to emerge and significantly influence the results. 
The probability of such event increases with decreasing number of objects; the effect is also boosted by overall large number of attributes, which in addition increases chances for random interactions between features.
This issue can be handled with by introducing artificial, random `contrast variables' \cite{Stoppiglia2003,Tuv2006,Rudnicki2006}, which are then used as a reference.

\subsection{Boruta}

Based on those observations, we have developed an alternative heuristics algorithm for \allrelevant feature selection --- Boruta \cite{Rudnicki2006,Kursa2010a,Kursa2010c}. 
It evaluates the relevance of variables in the information system by comparing the importance measure provided by \rf for the original attributes with that obtained for the artificially added random attributes. 
To this end it trains random forest classifier on the dataset extended with random contrast attributes, obtained by randomly shuffling values of original attributes between objects.  
Then, for each attribute, it is tested whether its importance is higher than the maximal importance achieved by any contrast attribute.
In order to obtain statistically significant results this procedure is repeated several times, with contrast variables generated independently for each iteration.
Boruta algorithm is implemented as an R \cite{R} package, and is available from CRAN. 

\subsection{Artificial Contrasts with Ensembles}

Another algorithm that can be used for the all-relevant feature selection is the modification of the ACE procedure proposed by Tuv and co-workers \cite{Tuv2006,Tuv2009a}.  
The general idea of the ACE algorithm is similar to the principle used by Boruta --- in order to establish absolute importance of the attribute the algorithm compares the importance of a real attribute with that of artificially constructed random features.
However, in contrast to Boruta, ACE is not removing found irrelevant variables to boost quality of importance measure but removes the effects of found important variables to allow more subtle interactions to appear.
Also the details of testing relevance are different.
The algorithm used here is an implementation of the original ACE procedure described in \cite{Tuv2006}, yet adopted to use the same source of importance measure as Brouta, namely using standard \rf tree growing procedure and larger forests (500 trees instead of 10).

\section{Related work}
Bayesian network inference is often performed as a wrapper over Na\"{\i}ve Bayes classifiers and could be used for all-relevant feature selection. 
However, since in all practical implementations the search for simple and previously postulated forms of node-node interactions \cite{Nilsson2007,Nilsson2007a,DeMorais2010}, these methods are not suitable for finding the non-trivial attributes, what is a subject of this study. 

The algorithm of Rogers \& Gunn \cite{Rogers2006} uses internals of \rf construction for feature selection. 
It relies on a theoretical model giving an estimate of the information gain of a split done on a non-informative attribute, which, averaged over the forest, is used to test the relevance of original attributes.
This method, while elegant, is not particularly good at discerning between relevant and irrelevant features. 
Authors present the results for the Madalon problem, where 130 features where deemed relevant by this algorithm at confidence level $p=0.001$, whereas there are only 20 relevant features in this set. 
For a comparison, this problem is solved nearly perfectly by Boruta algorithm \cite{Kursa2010c}.

The algorithm very similar to the first version of the Boruta \cite{Rudnicki2006} was recently proposed by Huynh-Thu et al. \cite{Huynh-Thu2008}.
It starts by obtaining the ranking of attributes from a single run of the \rf algorithm. 
Then, numerical simulations are used to estimate probability of achieving each importance level by a random fluctuation --- to this end, for each importance threshold, the attributes with higher importance are left unchanged, whereas the values of attributes with lower importance are permuted between objects (all attributes are permuted using identical permutation vector).  
Typically about 1000 iterations are required to establish a good estimate of the probability at each level due to the high variance of the importance score of randomised variables. 
The procedure is repeated for each decreasing importance level until the probability of achieving given level randomly reaches the predefined threshold of statistical significance. 

The Huynh-Thu algorithm is impractical for systems with large number of relevant attributes due to prohibitively large computational cost. 
For example, the system described with 10 relevant attributes requires at least 11 thousand repetitions of random forest.
This computational effort is more than two orders of magnitude larger than that required by Boruta, and the performance gap grows linearly with the number of relevant attributes in the system. 
Some optimisations of the proposed protocol are possible --- for instance the efficient search threshold value instead of linear scan --- but even then several thousands of RF runs are required. 
The extremely high computational load of the this algorithms was the reason of excluding it from the extensive tests performed in this study. 

\section{Experimental setup and methodology}
Three experiments have been performed.
The goal of the first one was to establish the usability of the \rf importance measure alone and as a base of two wrapper algorithms solving the \allrelevant feature selection problem.
In this test purely synthetic datasets were used, hence the real relevance of attributes was known a priori.

Next benchmark was devoted to analyse how the nature of the examined problem influences the performance of a heuristic algorithm.
The used here semi-synthetic datasets were created by extending four real-world low dimensional problems with random attributes, so that both the stability of the selection and the false positive rate could be estimated.

Finally the heuristic algorithm was used to analyse the  well-known experimental data set, the gene expression levels of leukaemia patients \cite{Golub1999h}. 
Only heuristic algorithms could be used for two latter tasks, since the number of relevant attributes is not known a priori in for the real-world data.
The results of the first test revealed that the computational efficiency of the ACE algorithm degrades rapidly for the datasets with large number of features, moreover the estimates of importance returned by ACE and Boruta were very similar, with better results achieved by Boruta.
Thus only Boruta heuristic algorithm was used for tests in this phase. 

\subsection{Synthetic data}
The first experiment was performed on a serious of synthetic geometrical problems analogous to the Madalon problem \cite{Guyon2005}, constructed with the help of \code{mlbench.xor} function from the R package \pkgname{mlbench} \cite{Leisch2010}.

Each of sets used consisted of two attributes used to generate the decision,
eight random linear combinations of them and as many irrelevant attributes made of an uniform noise to top up to the given size --- this way in all cases there was 10 attributes important by design.
All possible variants of such a data set containing 125, 250, 500, 1000 and 2000 objects described with 125, 250, 500, 1000, 2000 and 4000 attributes were generated.
Finally in each case the set of relevant attributes was calculated with all three wrapper algorithms described above; 
as a reference also 10 top important attributes from a single \rf run were saved.
One should note that given a feature importance, the essence of the \allrelevant feature selection is finding either an importance cut-off or correspondingly the number of top-scoring features that separates relevant and irrelevant attributes. 
Therefore an algorithm which deems important top $N$ attributes from the \rf importance ranking, where $N$ is the known a priori number of relevant attributes, is a reference that shows the limit of information contained in the importance measure.   
The entire procedure was repeated fifteen times.

Each result was described by the following quantities, averaged over fifteen iterations: number of attributes correctly recognised as important (true positives, $\TP$), attributes incorrectly recognised as important (false positives, $\FP$) and attributes which where incorrectly not recognised as important (false negatives, $\FN$).  
The overall performance was assessed as an average F-score, $F=\frac{2 \Prec \times \Rec}{\Prec+\Rec}$, where   
$\Prec = \frac{\TP}{\TP+\FP}$ and $\Rec = \frac{\TP}{\TP+\FN}$.

\subsection{Semi-synthetic data}
The second experiment used four series of semi-synthetic data sets. 
Each set was constructed from a real-world data set by extending it with an artificial attributes made of an uniform random noise
to obtain respectively 125, 250, 500, 1000 and 2000 total attributes, similarly to the construction of the synthetic test sets.
Thus one knows that the added attributes are irrelevant by design and correspondingly that the relevant attributes may be only found among original attributes. 
The Boruta algorithm was used to find the relevant attributes in all cases; the whole procedure was repeated 15 times and the results were averaged.

As a base real-world sets the following four well-known data sets were used: German Credit \cite{Frank2010}, House Votes \cite{Leisch2010,Frank2010}, Ozone \cite{Leisch2010,Breiman1985} and Vehicle \cite{Leisch2010,Frank2010,Siebert1987}.  

\subsection{Gene expression data} 
The third experiment was designed to show an example of a real-world \allrelevant feature selection use, namely the analysis of the results of microarray experiment. 
Such data sets usually comprise relatively few objects (often less than one hundred), and are described with several thousands of attributes. 
Moreover, finding  genes that are relevant for the studied phenomena is usually the main goal of the experimental study. 

A well known problem of detecting the differences between two subtypes of leukemia  from a seminal work of Golub \cite{Golub1999h}  was examined.
This data was deeply investigated before, thus gives a chance to compare the results of Boruta with three previously published results obtained using various other methods.

Several variants of this set were analysed in the literature and are available in public repositories; in this work, the data set \code{golub} available from \pkgname{hopach} Bioconductor package \cite{Pollard} was used. 
It consists of the data on 38 patients described with expression levels of 3051 genes. 
The raw data was preprocessed there as described in \cite{Dudoit2002h}.

The gene relevance was assessed with Boruta algorithm based on the \rf consisting of increasing number of trees.  
The number of trees in the forest varied between 500 and 100 000. 
Each run was repeated 15 times. 
To assess the possibility of false positive hits the analysis was repeated using a semi-synthetic version of Golub set. 
In this set the original attributes were augmented with 1000 additional synthetic attributes constructed by  randomly selecting 1000 original attributes and permuting their values between objects. 

\section{Results}

\subsection{Synthetic data}
\begin{figure}[tb]
	\centering
	\includegraphics[width=0.795\textwidth]{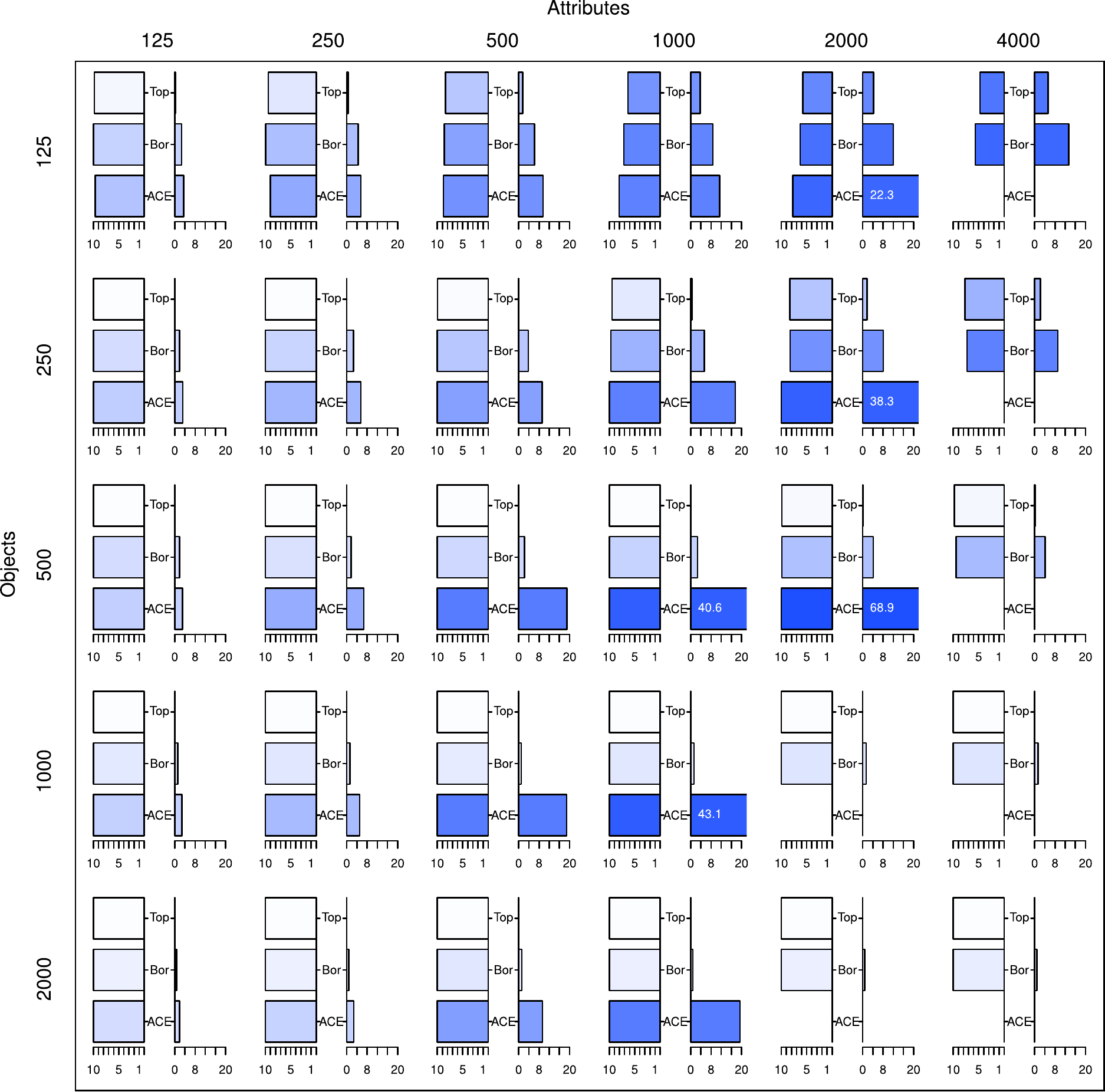}
	\caption{\label{fig:Pseudomadelon}
	Comparison of the \allrelevant feature selection algorithms on the artificial set. 
Each mini-plot represents averaged result of fifteen repetition of simulation performed for a system of size given by mini-plot's position in a table.
The length of bars on the left side of each mini plot are proportional to the number of properly identified relevant features.
The length of bars on the right are proportional to the number of false positive hits, unless that number is higher than twenty.
In such case the number of false positive hits is reported in the bar. 
The brightness of the bars is proportional to the F-score (white: F-score=1, dark blue: F-score=0). 
The results for the reference algorithm Top is on the top of each mini-plot, followed by Boruta and ACE.}
\end{figure}

The algorithms were compared for thirty parameter sets forming regular grid in the space of attribute number versus object number, see  Figure~\ref{fig:Pseudomadelon}.  
The problem difficulty grows with increasing number of attributes and decreasing number of objects, reaching maximum in the top right corner of the plot. 

All algorithms were using the same source of attribute importance. 
The reference algorithm Top returns best possible information retrievable from the single \rf run. 
It is used both to estimate quality of the \rf feature ranking and as a reference for the heuristic algorithms.  
For the remaining three algorithms the test was focused on the second issue of \allrelevant feature selection, the discerning between relevant and irrelevant features. 
Thus, it effectively tested if the method is able to discover the number of relevant features from the importance spectrum.

Three parameters were used to compare the results of four algorithms, namely the number of true positives (sensitivity), the number of  false positives (selectivity) and F-score which is a harmonic mean of two previous numbers and a good measure of the overall quality of the algorithm.  

\begin{figure}[ht]
	\centering
	\includegraphics[width=0.8\textwidth]{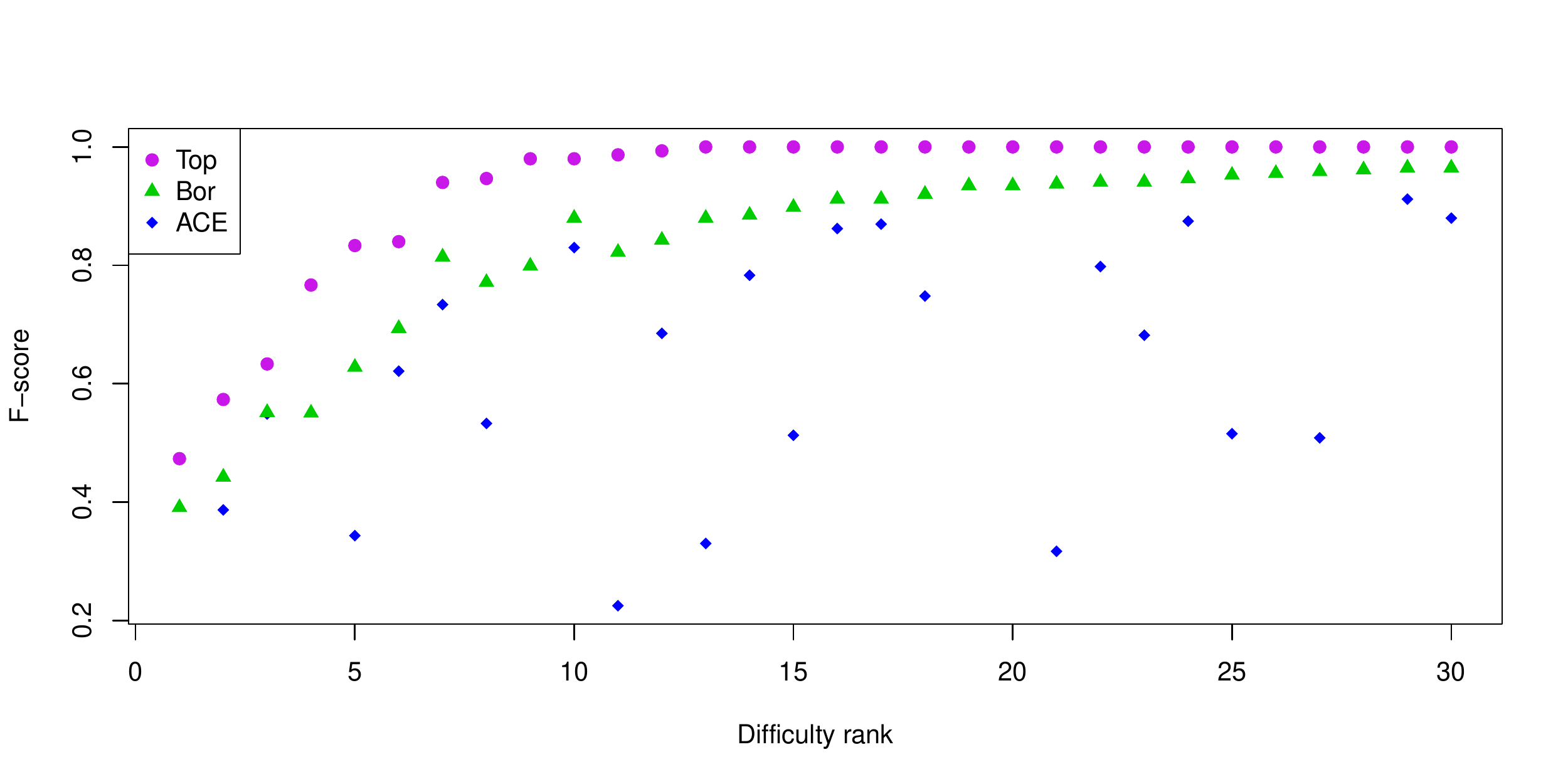}
	\caption{\label{fig:PseudomadelonFlat}
	The quality of the feature selection as a function of the problem difficulty. The difficulty rank is obtained by sorting the results first by the F-score obtained by Top algorithm and then by F-score obtained by Boruta.}
\end{figure}

\begin{figure}[ht]
	\centering
	\includegraphics[width=0.8\textwidth]{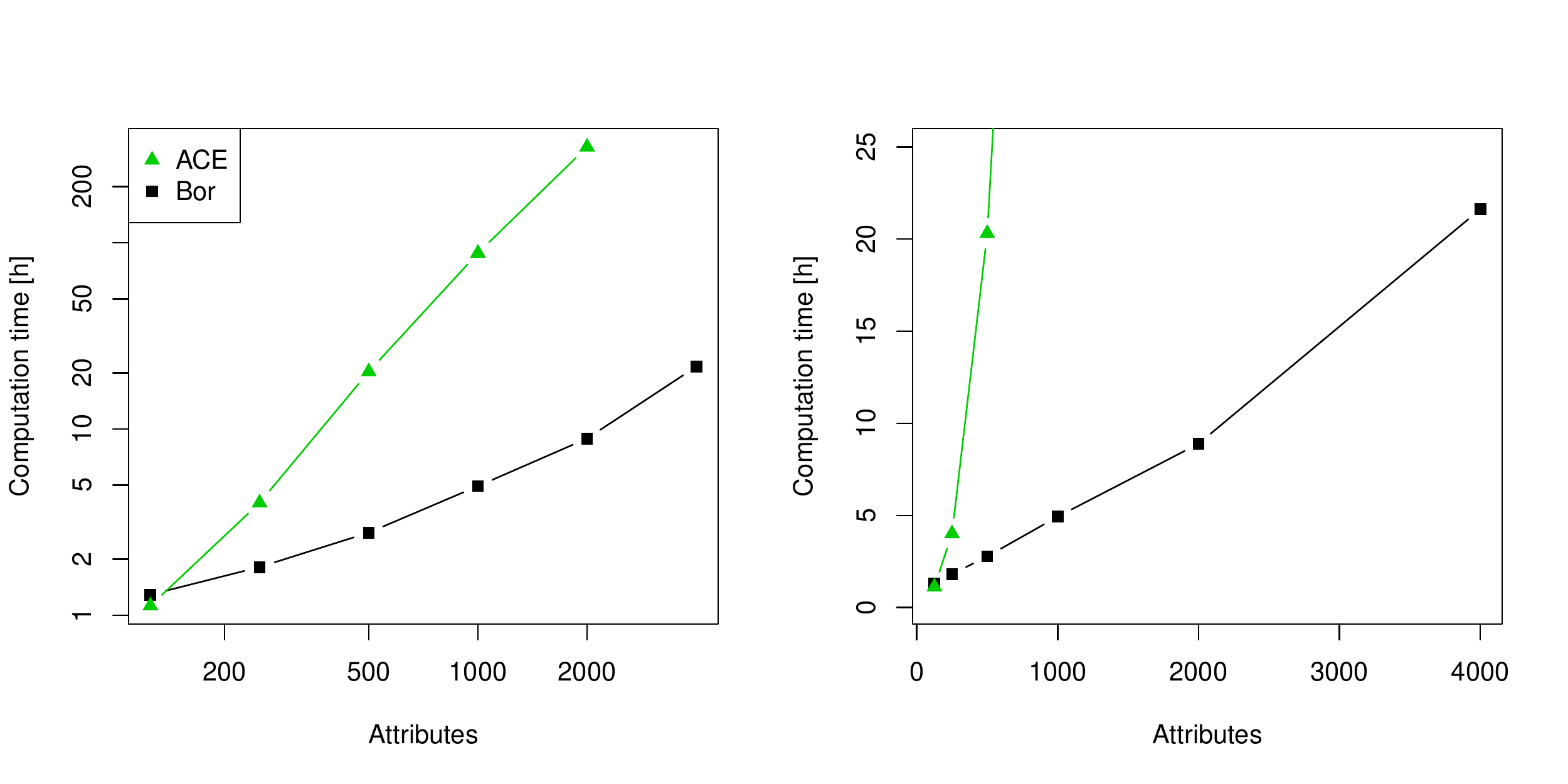}
	\caption{\label{fig:PseudomadelonTimes}
	Execution time of ACE and Boruta algorithms applied for system containing 500 objects. Logarithmic scale on the left panel is introduced to show all timings in the single plot.}
\end{figure}

As expected the F-score of reference algorithm Top was highest in all thirty cases. 
In eighteen cases it recognized all relevant attributes in all fifteen repetitions performed for given combination of number of attributes and number of objects. 
In four additional cases it identified correctly all relevant attributes in most iterations. 
The lowest score (F=0.47) was achieved for the system consisting of 125 objects and described with 2000 attributes, the highest (F=1) was achieved in 18 cases.

The ACE algorithm has a very good sensitivity. 
It was able to find all relevant attributes in 18 cases (19 in most iterations), even in the difficult cases. 
Unfortunately the execution time of this algorithm scales very badly with the number of attributes and algorithm did not converge in the available time (350 CPU hours) in 7 cases. 
These results suggest, that ACE algorithm is not suitable for systems described with very large number of features. 

The Boruta algorithm is overall better than ACE.
It has found all relevant attributes in 20 cases (23 in most iterations). 
Its sensitivity was slightly lower than ACE algorithms in few cases, but on the other hand much better selectivity. 
However, it achieved higher F-score than ACE algorithm in all cases, see Figure~\ref{fig:PseudomadelonFlat}.  
The highest score (F=0.93) was obtained for the system described with 125 attributes and 2000 objects.

All algorithms take substantial time to complete for large data sets, see Figure~\ref{fig:PseudomadelonTimes}.

The ACE algorithm is marginally faster than Boruta for low-dimensional sets, but the execution time increases steeply with the number of attributes. 
Already for 500 attributes ACE algorithm took roughly 10 times longer to complete, and it has not completed computations within 350 hours for any system described with 2000 attributes, regardless of the number of objects.  
The time requirements of the ACE algorithm make it impractical for systems described with large number of attributes. 

The results obtained in this section suggest that random forest is in most cases a reliable source of feature ranking.
Boruta algorithm, has two advantages over the ACE algorithm that make it more suitable for the all-relevant search.
It consistently scored highest in the F-score measure, its execution time is reasonable and the sensitivity is only slightly lower in some cases than that of ACE.

One should note, that this data set is a very tough test for a random forest importance ranking, due to the way the trees are constructed and how the importance is measured.
In the XOR problem the original attribute is informative only in conjunction with other original attribute and so they are rarely included in the trees, and rarely have high impact on the performance of the individual tree.  
The linear combinations of original attributes are much more useful.
This is reflected in the importance ranking returned by the forest, where the original attributes have low scores that may be lower than scores of random attributes in the case of data sets with few objects.  
This effect is decreasing for the larger sets. One possible reason is the decrease of the apparent importance of the random attributes with increasing number of objects, due to decreasing influence of random fluctuations. 
Another possibility is that the growing tree size increases the chance that at least two truly informative attributes are used for a tree construction, and therefore increases the importance of the cooperative attributes.  

\subsection{Semi-synthetic data}
\begin{figure}[t!]
	\centering
	\includegraphics[width=1.00\textwidth]{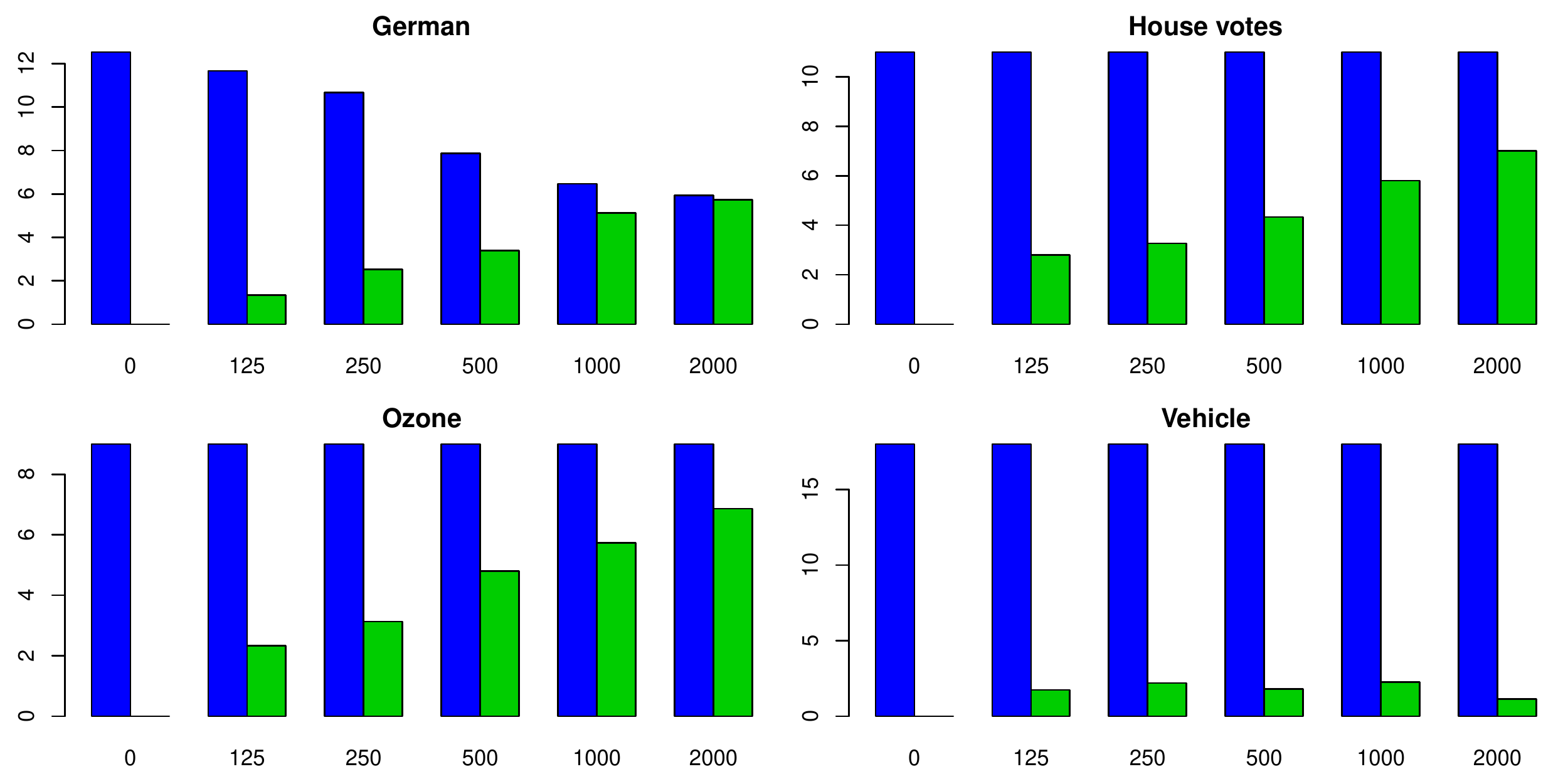}
	\caption{
	The results of Boruta algorithm applied to four families of semi-synthetic sets. The blue bars represent number of attributes that were deemed relevant for the ancestor set and were also deemed relevant for descendent sets. The green bars represent certain false positive hits, namely artificially added random attributes claimed relevant. }
	\label{fig:Subsynthetic}
\end{figure}

The experiment performed on four families of semi-synthetic data sets reveals some general tendencies, nevertheless it shows also that results for any data set may be significantly influenced by properties of the particular set, see Figure~\ref{fig:Subsynthetic}. 
In three families of semi-synthetic data  Boruta algorithm finds all attributes deemed important in the original data set in all extended sets, but in one set (German credit) the number of retrieved attributes declines steadily with increasing number of added variables.  
Interestingly, in one case (House votes) the attributes that were not deemed important in the original data set gained importance in extended data set (data not shown); this was not very strong effect (two new attributes compared with the eleven original ones), but consistent across all repetitions. 
In three families the number of false positive hits returned by Boruta was a slowly increasing function of number of attributes, starting from roughly 2 for system described with 125 attributes and reaching 6 for systems described with 2000 of attributes. 
In these three cases doubling of total number of attributes adds roughly one additional false positive attribute. 
This association does not turn up in the case of Vehicle data set, where the number of false positive hits is roughly constant for all set sizes. 

The examples studied in this test show that in most cases the sensitivity of the algorithm does not depend on the number of irrelevant attributes in the system.   
On the other hand the selectivity of the algorithm is decreasing with number of attributes, although this decrease is rather slow. 
One should note that the high variance of the artificial attributes, which were drawn from the uniform distribution, facilitates creation of fluctuations that may, in conjunction with other attributes, lead to creation of false rules. 
The observed tendencies should be considered a worst-case scenario.
As demonstrated in the next section, these effects are smaller or even non-existent, when the artificial variables are drawn from more realistic distributions.

\subsection{Gene expression data}
\begin{figure}[ht]
	\centering
	\includegraphics[width=0.80\textwidth]{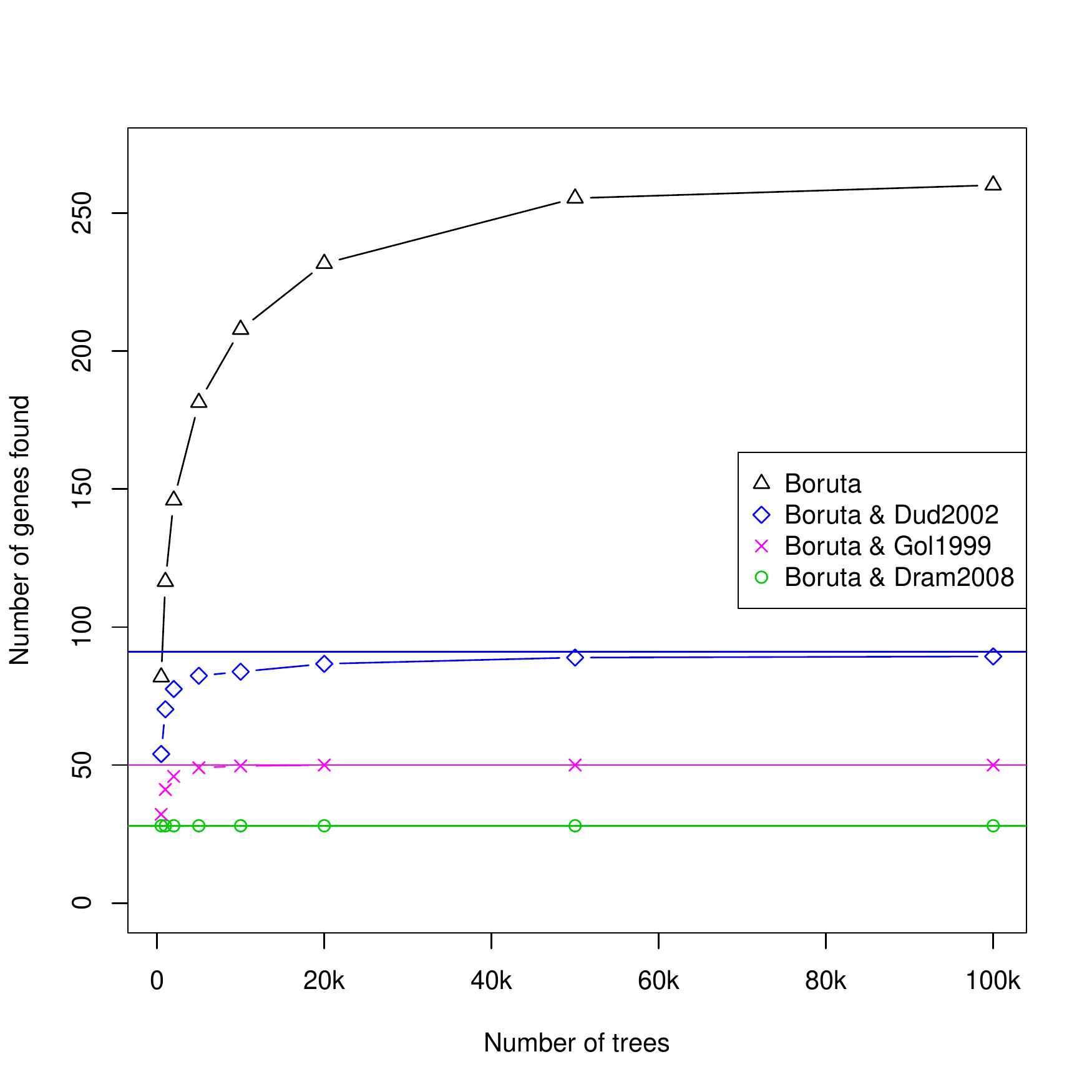}
	\caption{
	Number of relevant genes found by Boruta for different number of trees (black points).
	Additionally, the magenta, blue and green points represent the number of genes both found by Boruta and present respectively in Gol1999, Dud2002 and Dram2008 set.
	Accompanying solid lines show the total number of genes in those sets. 
	}
	\label{fig:AffyConvergence}
\end{figure}

The first major observation driven from the results of gene expression experiment is that the sensitivity of the algorithm significantly increases with the increasing number of trees grown in the random forest, see Figure~\ref{fig:AffyConvergence}. 
The number of important attributes grows logarithmically with increasing number of trees in the forest.
It starts from 82 relevant attributes found for the default forest size (500 trees). 
Then about twenty five additional relevant attributes are identified per doubling of the number of trees, up to fifty thousand trees. 
The last doubling, from fifty to one hundred thousand trees brings in only 5 new attributes.
The parallel control experiment was performed using the semi-synthetic version of the Golub set, where the original attributes were extended  with one thousand artificial contrast variables. 
In all fifteen repetitions of the control experiment not even a single contrast variable was ever deemed relevant by the algorithm, regardless of the number of trees in the random forest.
This is very strong hint that the attributes that were deemed relevant are indeed connected with decision. 
This is remarkable result when compared with results obtained both with the synthetic and the semi-synthetic data sets, where the non-negligible number of false positives hits were obtained. 

Yet the main aim of this experiment was to compare the results obtained with heuristic procedure based on random forests with the previously published lists of important genes. 
In the original paper Golub et al. reported 50 differentially over-expressed genes (further referred to as Gol1999). 
Dudoit et al. have reported 91 over-expressed genes (further referred to as Dud2002) and more recently the ranking of 30 genes, based on importance for classification by an ensemble of trees, was reported by Draminski et al. \cite{Draminski2008} (further referred to as Dram2008).
The data for the latter study was preprocessed differently than that used to obtain Dud2002, and in effect the set of important genes includes two genes which were not present neither in the gene set examined by Dudoit et al. \cite{Dudoit2002h} and in the current work. 
These genes were excluded from further analysis and therefore the Dram2008 set comprises 28 genes. 
The set of relevant attributes obtained with Boruta based on $N$ trees is further referred to as Bor$N$. 
Boruta500 set contains 82 genes, whereas the Boruta100k contains 261 genes. 

\begin{figure}[tb]
	\centering
	\includegraphics[width=0.8\textwidth]{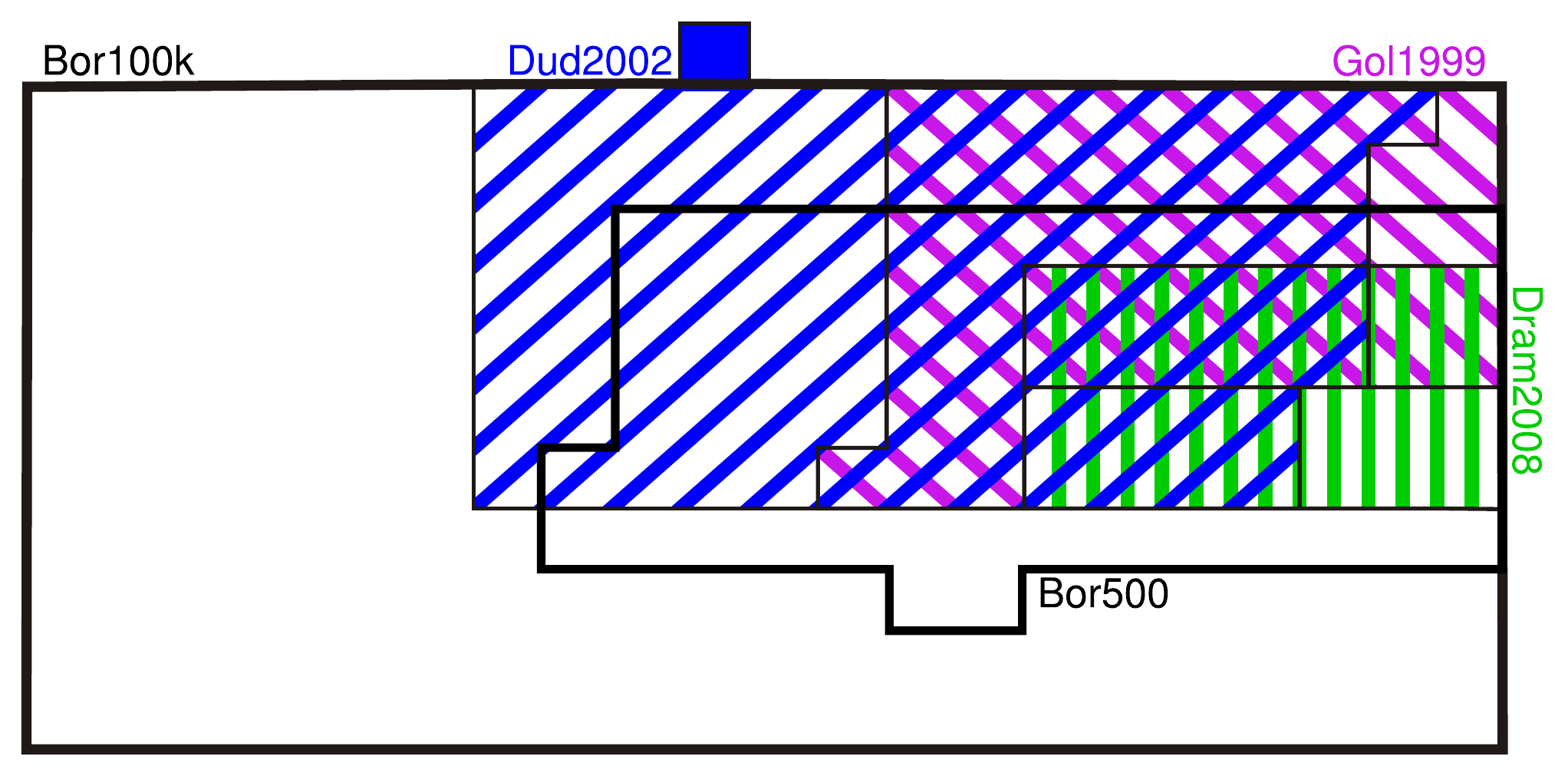}
	\caption{
	Graphic representation of the gene sets returned by various feature selection algorithms. The area covered with blue, magenta and green stripes represent intersection of the Bor100k with Dud2002, Gol1999 and Dram2008 data sets, respectively. The solid magenta square represents a single gene from Dud2002 that was not found by Boruta. The area encircled by an external solid line genes in the Bor100k set, the area encompassed within the internal solid line represents Bor500 set.}
	\label{fig:AffyAttmap}
\end{figure}

The comparison of genes deemed important by four methods is presented in Figure~\ref{fig:AffyAttmap}.  
One can easily notice, that all results of three older algorithms differ significantly. 
Five genes were unique to Gol1999, six to Dram2008 and 41 to Dud2002. 
Ten attributes are found in all sets,  four are common present in Gol1999 and Dram2008 but not in Dud2002, thirty one are present in Gol1999 and Dud2002 but not in Dram2008. 

The Bor500 set contains all genes from Dram2002, including six genes unique to this set. It contains also two genes unique to Gol1999,  twenty genes unique to Dud2002, as well as sixteen genes contained both in Gol1999 and Dud2002. 
There are sixteen genes unique to Bor500.
Interestingly, two algorithms using tree-based importance agree very well despite significant differences in the implementation and importance definition.  

The Bor100k set, as well as Bor50k set, contain all attributes found in all other sets, with the exception of a single gene present uniquely in Dud2002; additionally more than 150 new genes were found in Bor100k. 
Despite all differences between results of previous studies, results of Boruta procedure confirms that the genes found with the older methods are indeed relevant. 
On the other hand this agreement confirm the robustness of the Boruta algorithm.  
The biological analysis of the new genes found by Boruta is beyond the scope of the current work. 

The results of this experiment show that the sensitivity of the Boruta algorithm depends on the number of trees in the \rf ensemble. 
This happens due to the properties of the importance measure --- it is estimated as an average decrease of  accuracy of trees that use given attribute after the information on the value of this attribute is removed, therefore the relevance of a variable can be observed only when sufficiently large number of trees were built using information contained in it. 
In the case of weakly relevant attributes in the system with large number of relevant attributes, the fraction of trees that include given attribute during in the early stage of construction may be tiny, so a large number of trees is required before the relevance of such variables becomes significant. 

The heuristic Boruta algorithm applied to the semi-synthetic version of the Golub data set returned results that were very close to those obtained for the original set, with slightly lower number of important genes identified in the extended set for each number of trees in the \rf ensemble. 
This happens due to slightly lower quality of the importance measure the extended data set. 

\section{Conclusions}
The results of the the current study show that variable importance measure provided by the \rf classifier is a very useful base for the wrapper algorithms solving \allrelevant problem. 
The family of synthetic data sets used to test its utility presents a very difficult challenge for the tree-based classifier. 
Nevertheless, in most cases all of the relevant features were placed above the irrelevant ones in the \rf importance ranking.     

The heuristic methods based on the artificial contrast can efficiently use the ranking from the \rf to discern truly relevant attributes from irrelevant ones. 
The results of the heuristic procedure are only slightly worse than ideally possible to obtain with underlying feature ranking. 
The tests performed both on the synthetic and semi-synthetic sets shows that heuristic algorithm can generate false positive hits, and the number of them may be correlated with the number of attributes. 
On the other hand, it did not return any obvious false positive in the tests performed on the Golub data set, whereas it both confirmed independently the importance of the attributes identified by alternative methods and additionally identified many new important attributes.  
This result suggests that the sensitivity of the heuristic based on the artificial contrast to false positive hits depends on the properties of irrelevant attributes. 

The heuristic procedures employed here were developed  for the underlying \rf classifier, but it is general enough for use with any classifier which can return an importance ranking of attributes. 
Nevertheless, the \rf seems very well suited to the role of the classifier providing feature ranking for the \allrelevant feature selection algorithm because, due to construction of its importance measure, it is sensitive even to weakly relevant attributes.

\subsection*{Acknowledgements}
Computations were performed at ICM, Grant G34-5.

\bibliographystyle{splncs03}
\bibliography{Remlr}

\end{document}